\documentclass[runningheads]{llncs}
\usepackage[T1]{fontenc}
\usepackage{graphicx}
\usepackage{amssymb}
\usepackage{dsfont}
\usepackage{epstopdf}
\usepackage{listings}
\usepackage{amsmath}
\usepackage{multirow}
\usepackage{soul}
\usepackage{booktabs}
\usepackage{bbding}
\usepackage{makecell}
\usepackage{stfloats}
\usepackage{makecell}
\usepackage{xspace}
\usepackage{url}
\usepackage{wrapfig}
\usepackage{multirow}
\usepackage{makecell}

\usepackage[misc]{ifsym} 
\def\ie{\emph{i.e.}}

\def\sota{{\em state-of-the-art~}}

\usepackage{color}
\usepackage[table, dvipsnames]{xcolor}
\usepackage{graphicx}
\usepackage{amsmath}
\usepackage{booktabs} 
\usepackage{multirow}
\usepackage{tikz}

\definecolor{citecolor}{RGB}{34,139,34}
\definecolor{blue_REI}{HTML}{5a8ad2}
\definecolor{red_AZUKA}{HTML}{C55A11}
\definecolor{purple_SHINJI}{HTML}{C686EE}
\definecolor{mygray}{HTML}{E8E8E8}
\definecolor{teaserred}{HTML}{C14D4D}
\definecolor{hr}{HTML}{FFC28F}
\definecolor{lr}{HTML}{ABCBE3}

\usepackage[colorlinks=true,citecolor=green,urlcolor=purple_SHINJI]{hyperref}
	\author{Haipeng (Rydeen) Zhou\inst{1} \and
		Sicheng Yang\inst{2}\and
		Sihan Yang\inst{2}\and \\
		Jing Qin\inst{3} \and 
		Lei Chen\inst{1,4} \and
		Lei Zhu\inst{1,4}\textsuperscript{(\Letter)}}
	\authorrunning{H.Zhou et al.}
	
	\institute{The Hong Kong University of Science and Technology (Guangzhou)
		\and
		Xi'an Jiaotong University
		\and
		The Hong Kong Polytechnic University
		\and
		The Hong Kong University of Science and Technology
	}

\begin{document}
	\title{CoC: Chain-of-Cancer based on Cross-Modal Autoregressive Traction for Survival Prediction}
	\titlerunning{CoC: Chain-of-Cancer for Survival Prediction}
	\maketitle
	
	\renewcommand{\thefootnote}{}

\begin{abstract}
Survival prediction aims to evaluate the risk level of cancer patients. Existing methods primarily rely on pathology and genomics data, either individually or in combination.
From the perspective of cancer pathogenesis, epigenetic changes, such as methylation data, could also be crucial for this task. Furthermore, no previous endeavors have utilized textual descriptions to guide the prediction.
To this end, we are the first to explore the use of four modalities, including three clinical modalities and language, for conducting survival prediction.
In detail, we are motivated by the Chain-of-Thought (CoT) to propose the Chain-of-Cancer (CoC) framework, focusing on intra-learning and inter-learning. We encode the clinical data as the raw features, which remain domain-specific knowledge for intra-learning. In terms of inter-learning, we use language to prompt the raw features and introduce an Autoregressive Mutual Traction module for synergistic representation. This tailored framework facilitates joint learning among multiple modalities.
Our approach is evaluated across five public cancer datasets, and extensive experiments validate the effectiveness of our methods and proposed designs, leading to producing \sota results.
Codes will be released\textsuperscript{\textcolor{purple_SHINJI}{1}}~\footnote{\textsuperscript{\textcolor{purple_SHINJI}{1:}}\url{https://github.com/haipengzhou856/CoC}}.
\end{abstract}
	
\section{Introduction}
Survival prediction~\cite{cox1972regression,cox1975partial}, particularly in oncology, plays a pivotal role in guiding clinical decision-making and personalized treatment strategies. It utilizes clinical data as biomarkers, aiming to provide risk stratification for patients. Recently, the use of data in survival analysis has primarily focused on genes and pathological images, leading to the development of both single-modality~\cite{klambauer2017self,zaheer2017deep,ilse2018attention,lu2021data,shao2021transmil} and multimodal learning methods~\cite{chen2021multimodal,zhou2023cross,xu2023multimodal,jaume2024modeling,xiong2024mome} that integrate the two.

Relying solely on genomic data poses challenges due to the high dimensionality of genomic profiles~\cite{qiu2020meta}. Consequently, single-modality approaches often adopt Multiple Instance Learning (MIL)~\cite{amores2013multiple} for processing Whole Slide Images (WSIs), \textit{i.e.}, pathology images. By dividing the giga-pixel resolution of WSIs into smaller patches, these models can focus on learning specific patterns from regions of interest. 
However, recent advances in combining genomic data with WSIs have demonstrated improved performance. These two modalities provide complementary information, enabling a more comprehensive understanding of the disease. For example, SurvPath~\cite{jaume2024modeling} suggests that tumor morphologies correspond well to the pathways in transomics. Specific genes influence the morphological features observed in pathology, and their synergistic relationship can be effectively uncovered through multimodal learning.

%, in which the WSIs will be cropped patch by patch and the model learn to build up a 
%However, the cancer-related 
%\textbf{why using language and meth?}
%In vision-end, we focus on the computational xxx.  As pathologists' diagnostic process, we follow the global-local fashion.

%In the omic-end, these tabular data possess very high-dimension.

%However, survival analysis relying solely on a single modality is far from sufficient.

%The genomics data is complementary to the expression of pathological images in survival analysis~\cite{hoang2024deep}, and several deep models~\cite{shao2021transmil,zhou2023cross,xu2023multimodal,jaume2024modeling,xiong2024mome} have explored the usage of WSIs and gene data. \textbf{TODO: for example, xxx. These are in common}.
Nevertheless, epigenetics (e.g., methylation data)\cite{verma2002epigenetics} also plays a vital role in regulating gene expression, impacting various biological processes and resulting in specific pathological imaging features like increased cell density and abnormal cell morphology \cite{zheng2020whole} in WSIs. By integrating genomic, epigenetic, and WSI data, we can achieve a comprehensive understanding of diseases from molecular, epigenetic, and morphological perspectives. Furthermore, with advancements in Large Language Models (LLMs), an increasing number of downstream tasks \cite{radford2021learning,kojima2022large,lu2024avisionlanguage,saharia2022photorealistic,wang2023dynamic,wang2024language,zhou2024timeline,xing2024segmamba} are exploring textual guidance to enhance their outcomes. Inspired by this, we aim to leverage these multiple modalities to unleash their potential to improve survival prediction.

In this paper, we propose a cross-modal autoregressive model, named Chain-of-Cancer (CoC). Our core idea is to utilize task-specific handcrafted language descriptions as the initial prompt, encouraging mutual learning across different modalities in an autoregressive manner. 
In particular, \textbf{1) we introduce a CoC-Adapter.} By providing clinical-related descriptions, we can embed textual guidance into different modalities to enhance representation.
\textbf{2) Moreover, we propose an Autoregressive Mutual Traction (AMT) module} to facilitate synergistic learning. This module establishes dependencies among different modalities to enable cross-modal learning.
\textbf{3) We leverage the inter-learning and intra-learning for survival prediction}. We encode the clinical data to extract raw features for intra-learning, and we deploy the CoC-Adapter and AMT module to conduct inter-learning. Through their combined effects, we achieve promising results on this task.
To the best of our knowledge, \textbf{we are the first to apply these new modalities (\textit{i.e.}, methylation and language) to survival prediction.} The tailored autoregressive learning manner also empowers our model to achieve promising results. We evaluate our method on five public datasets, and the experimental results demonstrate that our CoC consistently surpasses \textit{state-of-the-art} methods.

\section{Method} 
\begin{figure*}[t!]
	\centering
	\includegraphics[width=1\linewidth, height=0.66\columnwidth]{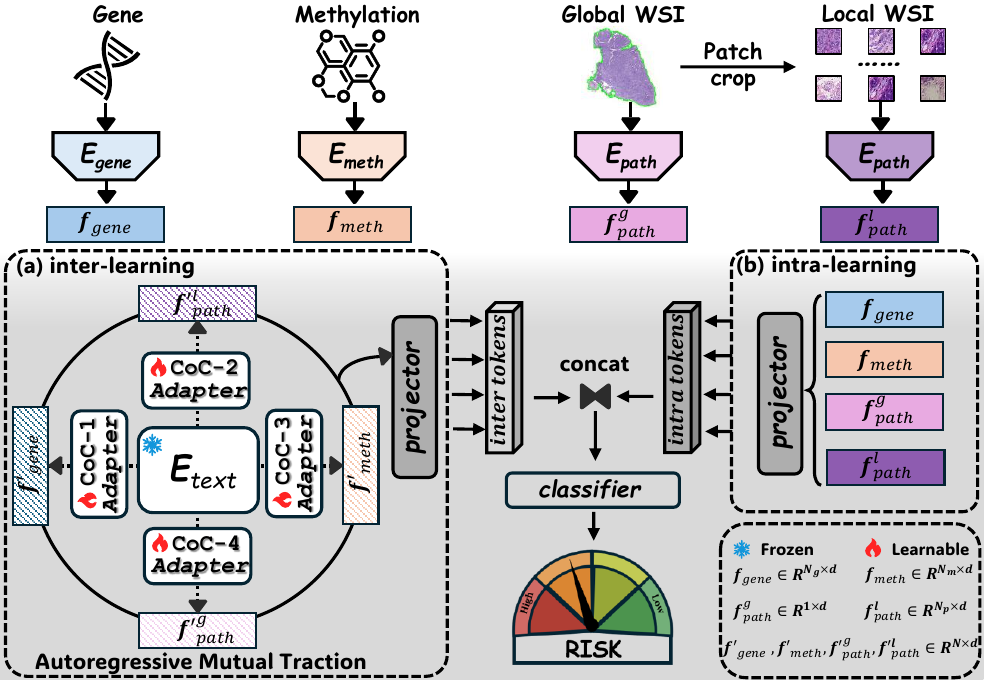}
	\caption{An overview of our CoC framework. It consists of the basic feature extraction, the (a) inter-learning, and the (b) intra-learning. In (a) we use CoC-Adapter to embed language guidance to produce homogeneity features autoregressively. In (b) we use a simple projector to yield heterogeneity features. Finally, by concatenating these features we can utilize a classifier to conduct survival prediction.}
	\label{fig1}
\end{figure*}
\subsection{Formulation and Overview}
Given the clinical data $\mathbb{X}$ and time-to-event label $\mathbb{Y}=\{t, c\}$, where $t\in \mathcal{R^{+}}$ is the overall survival time and $c\in \{0,1\}$ is event censorship at $t$, our target is to estimate the death probability via the hazard function $f(t) = f(T=t|T\geq t, \mathbb{X})$. Such that, we can estimate the ordinal risk of death event occurring at time point $t$ via a discrete-time format~\cite{haider2020effective} by $S(t|\mathbb{X}) = \prod^{t}_{j}(1-f(j) )$. The hazard-based task can be converted into classification~\cite{zadeh2020bias} by setting the number of time bins being $4$, and we have negative log-likelihood loss with the censorship $c$ termed as:
\begin{equation}
	\mathcal{L}_{surv} = -c~\mathrm{log}(S(t|\mathbb{X})).
\end{equation}

As shown in Fig.~\ref{fig1}, in our study we have $\mathbb{X} = \{ \mathcal{X}_{gene},\mathcal{X}_{meth}, \mathcal{X}_{path}^{g},\mathcal{X}_{path}^{l}\}$ which denotes the 1-D gene profile, and the 1-D methylation profile, the pathology images at global-level, and the cropped patches,  respectively. We first conduct the feature engineering to obtain their features yielding $f_{gene} \in \mathcal{R}^{N_g\times d}$, $f_{meth} \in \mathcal{R}^{N_m\times d}$, $f_{path}^{g} \in \mathcal{R}^{1\times d}$,  and $f_{path}^{l} \in \mathcal{R}^{N_p\times d}$, where $N_g$ and $N_m$ are the number of tokens for gene and methylation features, $N_p$ is the number of local patches, and $d$ is the latent dimension. These raw features are heterogeneous, and we simply retain them and feed them into a projector to conduct intra-learning. In terms of inter-learning, we explicitly use descriptive text to embed these features into synergistic knowledge space via CoC-Adapter. Then, we deploy Autogregressive Mutual Tranction to continue mine the homogeneous features. By leverage intra-learning and inter-learning, our network can effectively explore the synergistic and independent contributions of each modality.

%Given the  pathology images at global-level $\mathcal{X}_{path}^{g}$, the cropped patches $\mathcal{X}_{path}^{l}$, the 1-D gene profile $\mathcal{X}_{gene}$, and the 1-D methylation profile $\mathcal{X}_{meth}$, our target is to predict the time-discrete label by $\hat{Y}=\{t, \delta\}$ where probability of death for a patient right after the time point t
\begin{figure*}[t!]
	\centering
	\includegraphics[width=1\linewidth]{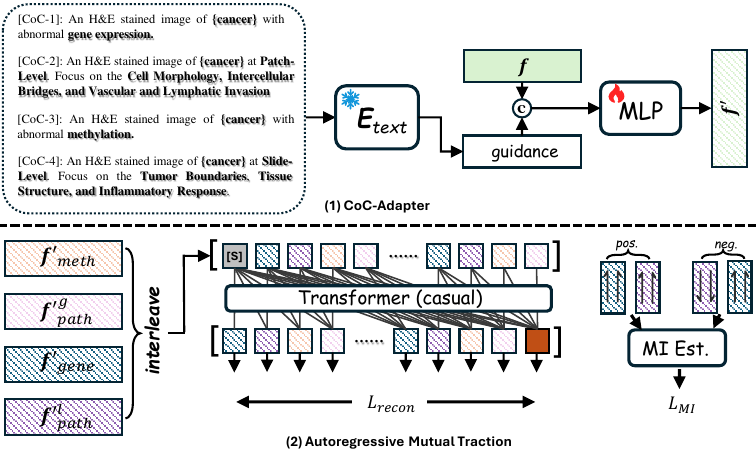}
	\caption{The Inter-Learning branch and it consists of two steps. (1) We first use CoC-Adapter to embed the raw feature with the text guidance. (2) We deploy Autoregressive Mutual Traction (AMT) module to conduct synergistic representation.}
	\label{fig2}
\end{figure*}
\subsection{Chain-of-Cancer Adapter}
%The vision language models (VLM), such as CLIP~\cite{radford2021learning}, have presented promising results in various tasks. Pretrained with text-image pairs, the CLIP-based methods enable providing concrete focus on the image via the specific linguistic description~\cite{zeng2024peeling}. 
Prompt methods like Chain-of-Thought (CoT)~\cite{wei2022chain,kojima2022large,zheng2023ddcot} demonstrate that explicitly using language prompts can enhance the model's reasoning ability for multimodal learning. Motivated by it, we propose the Chain-of-Cancer Adapter, which makes use of clinical-related descriptions as textual guidance to integrate with different modalities.

Fig.~\ref{fig2} (1) shows the workflow of our CoC-Adapter. We first design tailored descriptions. For all the modalities, we give a vanilla prompt "\texttt{An H\&E stained image of {\textbf{\{cancer\}}}}" where `cancer' denotes the cancer type. Considering the specific knowledge, we detail the prompt according to the modality. For the visual end, we analogize the pathologists' diagnosis in which we provide the priors from global and local perspectives, \textit{e.g.}, the Tumor Boundaries \textit{vs.} Intercellular Bridges. In terms of the 1-D data, we add the suffix class (\textit{e.g.}, `methylation') to inject priors. Thus, we can deploy a text encoder $E_{text}$ to generate language guidance embedding. Then, we concatenate the raw feature $f$ with the guidance and feed it into a learnable MLP layer, yielding the text-embedded feature $f'$.
%\begin{equation}
%	f' = \mathrm{MLP}(\mathrm{concat}(E_{text}(\mathrm{prompt}),f) )
%\end{equation}

\subsection{Autoregressive Mutual Traction (AMT)}
We first introduce the preliminary of autoregressive models (ARM). Given a sequence of tokens $x=\{x^1, x^2,..., x^n\}$, ARM predict the current token $x_i$ based on previous tokens $\{x^{j}\}_{j\leq i-1}$, and the next token prediction can be termed as:
\begin{equation}
	\label{eq1}
	p(x^1, ...,x^n) = \prod_{i=1}^{n}p(x^i|x^1, ..., x^{i-1}).
\end{equation}
Thus, an ARM parameterized by $\theta$ is to optimize the probability $p_{\theta}(x^i|x^1, ..., x^{i-1})$.

We propose that the autoregressive structure facilitates cross-modal interaction and homogeneity representation by enforcing causal dependencies between different modalities through the next-step prediction paradigm. As shown in Fig.~\ref{fig2}, in our approach, we interleave the features to form a sequence $x$, which helps prevent the model from focusing exclusively on features within a single modality. Following previous ARM~\cite{tian2024visual,li2024autoregressive}, we add a start token [\texttt{s}] to serve as the context for predicting the first element in the sequence. Such that, we can deploy a standard Transformer with 2 layers to decode the reconstruction. To optimize it, we have:
\begin{equation}
\mathcal{L}_{rec} = ||x - \hat{x}|| ^{2}.
\end{equation}

Besides, to prevent over-reconstruction we design a Mutual Information regulation to make the model retain meaningful cross-modal relationships. Given a pair of reconstructed features from different modalities, $m_1$ and $m_2$, we have:
\begin{equation}
	\mathcal{L}_{MI}^{m1, m2} = \sum_{(m_1, m_2)} -\mathbb{E}_{(x_i^{m_1}, x_i^{m_2}), (\tilde{x}_j^{m_2})} \left[ \log \sigma \left( f_{\phi}(x_i^{m_1}, x_i^{m_2}) - f_{\phi}(x_i^{m_1}, \tilde{x}_j^{m_2}) \right) \right],
\end{equation}
where $x_i^{m_2}$ is the positive samples predicted by the ARM. For negative samples $\tilde{x}_j^{m_2}$, since in ARM we generate them in an interleaved causal manner, we can use negative samples by randomly shuffling the order of this modality. The $\phi$ is a mutual information estimator which can be a simple MLP layer. Thus, considering the four traction chains (including the global and local pathology features), we will have a total of 6 (3$\times$2) pairs to compute the mutual information.

%We propose that .With the text-embedded features from CoC-Adapter \{$f'_{gene}$, $f'_{meth}$, $f'^{l}_{path}$, $f'^{g}_{path}$\}, 
\subsection{Objective}
For our intra-learning, we can concatenate the raw features and use a linear projector layer to out intra-tokens. Similarly, we also deploy a linear projection for the output of AMT yielding the inter-token. Considering the specific and synergistic knowledge jointly, we concatenate them and feed them into the classifier to predict the patient risk. The total term of the loss function is computed as:
\begin{equation}
	\mathcal{L} = \mathcal{L}_{surv} + \mathcal{L}_{rec}+ \lambda*\mathcal{L}_{MI}, 
\end{equation}
where $\lambda$ is empirically set as $0.3$.
\section{Experiments and Results}
\subsection{Datasets}
We use the public available data from TCGA\textsuperscript{\textcolor{purple_SHINJI}{1}}~\footnote{\textsuperscript{\textcolor{purple_SHINJI}{1:}}\url{https://portal.gdc.cancer.gov}}, including WSIs, genomic data, the methylation data, and the ground truth of survival time. A total of 5 cancer datasets are used,  including Cervical squamous cell carcinoma and endocervical adenocarcinoma (CESC, 270 cases), 	Liver hepatocellular carcinoma (LIHC, 350 cases), Breast invasive carcinoma (BRCA, 1058 cases), Colon adenocarcinoma (COAD, 444 cases), and Kidney renal clear cell carcinoma (KIRC, 509 cases). All the cases are cleaned with 60,660 genomics features and 80,000 methylation features.  The overall data volume is about 2.5 TB.

\subsection{Implementation Details}
\subsubsection{Metrics.}
Following previous works, we adopt C-Index~\cite{harrell1996multivariable} as the quantitative metric to evaluate the performance. We conduct 5-fold cross-validation to ensure the reproduction and robustness. The Kaplan-Meier analysis~\cite{syriopoulou2022standardised,kaplan1958nonparametric} is also presented, along with an evaluation of the log-rank test.

\subsubsection{Configurations.}
%All the experiments are conducted on a single Nvidia A6000 GPU based on Pytorch.
We use AdamW optimizer, a learning rate of 1e-4, and 20 epochs to train the model. In line with previous works\cite{zaheer2017deep,lu2021data,shao2021transmil,zhou2023cross,xu2023multimodal,jaume2024modeling,xiong2024mome,li2025towards,wang2024dual}, 
we utilize ResNet-50~\cite{he2016deep} to encode pathology data and use SNN~\cite{klambauer2017self} to encode 1-D data (\ie, gene and methylation). In terms of language-end, we deploy the tokenizer from CONCH~\cite{lu2024avisionlanguage} which is trained by WSI-Text pairs. This enables us to make use of handcrafted descriptive texts in a medical domain instead of vanilla CLIP~\cite{radford2021learning}. In our implementation, we have $N_{g}=6$, $N_{m}=8$, $N=4$, and $d=512$, respectively. And $N_p$ is determined by the WSI tiling resulting in various values. For other methods, we reproduce them by their official codes.

% instead of vanilla CLIP~\cite{radford2021learning} in order to reduce the domain gap. 

%WSIs is tiled with $256\times256$ patch size and encoded by ResNet-50~\cite{he2016deep} in line with previous works~\cite{zaheer2017deep,lu2021data,shao2021transmil,zhou2023cross,xu2023multimodal,jaume2024modeling,xiong2024mome}. 

%The embedding dimension is set with 1024.
\begin{table}[t]
	\caption{Quantitative comparisons with \textit{state-of-the-art} methods under the metric of C-index (mean $\pm$ std) on  5 cancer datasets with 5-fold cross-validation.
		%We compare a total of 13 approaches on 5 cancer datasets with 5-fold cross-validation.
		The best ones and runner-ups are highlighted with \textcolor{red_AZUKA}{\textbf{red}} and \textcolor{blue_REI}{\textbf{blue}}, respectively. The \includegraphics[height=10pt]{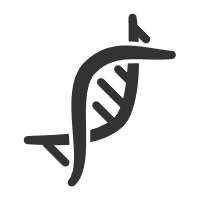}, \includegraphics[height=10pt]{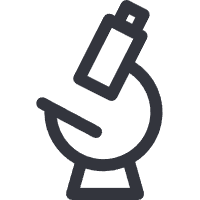}, and \includegraphics[height=10pt]{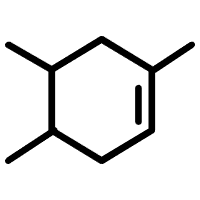} denote the modality of gene, pathology and methylation, respectively.}
	\renewcommand{\arraystretch}{1.1}
	\centering
	\resizebox{1\textwidth}{!}{
		\begin{tabular}{c|c|cccccc}
			\Xhline{3\arrayrulewidth}
			Methods & Modality & \multicolumn{1}{c}{\begin{tabular}[c]{@{}c@{}}CESC\\ (N=270)\end{tabular}} & \multicolumn{1}{c}{\begin{tabular}[c]{@{}c@{}}LIHC\\ (N=350)\end{tabular}} & \multicolumn{1}{c}{\begin{tabular}[c]{@{}c@{}}BRCA\\ (N=1058)\end{tabular}} & \multicolumn{1}{c}{\begin{tabular}[c]{@{}c@{}}COAD\\ (N=444)\end{tabular}} & \multicolumn{1}{c}{\begin{tabular}[c]{@{}c@{}}KIRC\\ (N=509)\end{tabular}} & \multicolumn{1}{c}{Overall} \\
			\Xhline{2\arrayrulewidth}
			MLP~\cite{bengio2000neural} & \includegraphics[height=10pt]{figs/gene.png} & $0.589_{\pm0.054}$ & $0.586_{\pm 0.044}$ & $0.621_{\pm 0.028}$ & $0.597_{\pm 0.025}$ & $0.655_{\pm0.059}$ & $0.610$ \\
			SNN~\cite{klambauer2017self} & \includegraphics[height=10pt]{figs/gene.png} & $0.573_{\pm0.052}$ & $0.591_{\pm0.037}$ & $0.622_{\pm 0.027}$ & $0.600_{\pm 0.018}$ & $0.664_{\pm0.052}$ & $0.610$ \\
			SNNTrans~\cite{klambauer2017self} & \includegraphics[height=10pt]{figs/gene.png} & $0.581_{\pm 0.046}$ & $0.600_{\pm 0.057}$ & $0.624_{\pm 0.012}$ & $0.602_{\pm 0.037}$ & $0.661_{\pm 0.043}$ & $0.614$ \\
			\hline
			SNN~\cite{klambauer2017self} & \includegraphics[height=10pt]{figs/meth.png}  &$0.611_{\pm 0.039}$&$0.594_{\pm 0.036}$&$0.594_{\pm 0.006}$&$0.588_{\pm0.037}$&$0.651_{\pm0.045}$& $0.608$\\
			SNNTrans~\cite{klambauer2017self} & \includegraphics[height=10pt]{figs/meth.png} &$0.609_{\pm 0.033}$&$0.596_{\pm 0.046}$&$0.611_{\pm0.034}$&$0.591_{\pm0.030}$&$0.653_{\pm0.031}$&$0.612$\\
			\hline
			Deep-sets~\cite{zaheer2017deep} & \includegraphics[height=9pt]{figs/path.png} & $0.541_{\pm 0.023}$ & $0.501_{\pm 0.003}$ & $0.541_{\pm 0.026}$ & $0.518_{\pm 0.031}$ & $0.512_{\pm 0.027}$ & $0.523$ \\
			AttnMIL~\cite{ilse2018attention} & \includegraphics[height=9pt]{figs/path.png} & $0.613_{\pm 0.065}$ & $0.505_{\pm 0.053}$ & $0.602_{\pm 0.014}$ & $0.574_{\pm 0.036}$ & $0.546_{\pm 0.035}$ & $0.568$ \\
			CLAM-SB~\cite{lu2021data} & \includegraphics[height=9pt]{figs/path.png} & $0.618_{\pm 0.066}$ & $0.526_{\pm 0.071}$ & $0.602_{\pm 0.015}$ & $0.569_{\pm 0.035}$ & $0.555_{\pm 0.036}$ & $0.574$ \\
			CLAM-MB~\cite{lu2021data} & \includegraphics[height=9pt]{figs/path.png} & $0.604_{\pm 0.073}$ & $0.508_{\pm 0.062}$ & $0.591_{\pm 0.014}$ & $0.567_{\pm 0.034}$ & $0.554_{\pm 0.035}$ & $0.565$ \\
			TransMIL~\cite{shao2021transmil} & \includegraphics[height=9pt]{figs/path.png} & $0.612_{\pm 0.044}$ & $0.611_{\pm 0.040}$ & $0.607_{\pm 0.017}$ & $0.595_{\pm 0.026}$ & $0.602_{\pm 0.009}$ & $0.605$ \\
			\hline
			SNN~\cite{klambauer2017self} & \includegraphics[height=9pt]{figs/gene.png}+\includegraphics[height=10pt]{figs/meth.png}  &$0.613_{\pm0.044}$&$0.590_{\pm0.057}$&$0.613_{\pm0.027}$&$0.604_{\pm0.020}$&$0.669_{\pm0.013}$&0.618 \\
			SNNTrans~\cite{klambauer2017self} &\includegraphics[height=9pt]{figs/gene.png}+ \includegraphics[height=10pt]{figs/meth.png} &$0.610_{\pm0.038}$&$0.603_{\pm0.041}$&$0.628_{\pm0.016}$&$0.606_{\pm0.029}$&$0.674_{\pm0.051}$& 0.624\\
			\hline
			%MCAT~\cite{chen2021multimodal} & \includegraphics[height=9pt]{figs/gene.png}+\includegraphics[height=9pt]{figs/path.png} & $0.629_{\pm 0.050}$ & $0.611_{\pm 0.040}$ & \textcolor{blue_REI}{$0.649_{\pm0.022}$}& $0.611_{\pm 0.024}$ & $0.682_{\pm 0.045}$ & $0.636$ \\
			MCAT~\cite{chen2021multimodal} & \includegraphics[height=9pt]{figs/gene.png}+\includegraphics[height=9pt]{figs/path.png} & $0.573_{\pm 0.034}$ & $0.604_{\pm 0.040}$ & $0.588_{\pm0.007}$& $0.590_{\pm 0.051}$ & $0.656_{\pm 0.017}$ & $0.602$ \\
			CMTA~\cite{zhou2023cross} & \includegraphics[height=9pt]{figs/gene.png}+\includegraphics[height=9pt]{figs/path.png} & $0.622_{\pm 0.061}$ & $0.595_{\pm 0.020}$ & $0.636_{\pm 0.013}$ & $0.605_{\pm 0.020}$ & $0.675_{\pm 0.029}$ & $0.627$ \\
			MOTCAT~\cite{xu2023multimodal} & \includegraphics[height=9pt]{figs/gene.png}+\includegraphics[height=9pt]{figs/path.png} & \textcolor{blue_REI}{0.637$_{\pm0.035}$} & $0.613_{\pm 0.028}$ & $0.630_{\pm 0.016}$ & $0.615_{\pm 0.016}$ & \textcolor{blue_REI}{0.696$_{\pm 0.054}$} & \textcolor{blue_REI}{${0.638}$} \\
			SurvPath~\cite{jaume2024modeling} & \includegraphics[height=9pt]{figs/gene.png}+\includegraphics[height=9pt]{figs/path.png} & $0.627_{\pm 0.044}$ & \textcolor{blue_REI}{0.616$_{\pm 0.047}$} & \textcolor{blue_REI}{$0.639_{\pm 0.032}$} & \textcolor{blue_REI}{0.618$_{\pm 0.017}$} & $0.685_{\pm 0.036}$ & $0.637$ \\
			MoME~\cite{xiong2024mome} & \includegraphics[height=9pt]{figs/gene.png}+\includegraphics[height=9pt]{figs/path.png} & $0.621_{\pm 0.037}$ & $0.610_{\pm 0.031}$ & $0.625_{\pm 0.049}$ & $0.610_{\pm 0.031}$ & $0.677_{\pm 0.038}$ & $0.628$ \\
			\hline
			\rowcolor{purple_SHINJI!8} 
			Ours & \includegraphics[height=9pt]{figs/gene.png}+\includegraphics[height=9pt]{figs/path.png}+\includegraphics[height=9pt]{figs/meth.png} &\textcolor{red_AZUKA}{\textbf{0.643}$_{\pm 0.028}$} &\textcolor{red_AZUKA}{\textbf{0.630}$_{\pm 0.047}$}&\textcolor{red_AZUKA}{\textbf{0.654}$_{\pm 0.036}$}&\textcolor{red_AZUKA}{\textbf{0.638}$_{\pm 0.046}$}&\textcolor{red_AZUKA}{\textbf{0.709}$_{\pm 0.048}$}& \textcolor{red_AZUKA}{\textbf{0.655}}\\
			\Xhline{2\arrayrulewidth}
		\end{tabular}	
		\label{table1}
	}
\end{table}
\subsection{Performance Evaluation}
\subsubsection{Comparisons with \sota Methods.}
We present the quantitative comparisons in Tab.~\ref{table1}, showcasing a total of 12 methods. For 1-D data, we utilize MLP~\cite{bengio2000neural}, SNN~\cite{klambauer2017self}, and SNNTrans~\cite{klambauer2017self} to generate single-modal and cross-modal results. For single pathology data, we reproduce Deep-sets~\cite{zaheer2017deep}, AttnMIL~\cite{ilse2018attention}, CLAM~\cite{lu2021data}, and TransMIL~\cite{shao2021transmil}. Additionally, we reconstruct dual-modal methods involving genomics and WSIs, such as MCAT~\cite{chen2021multimodal}, MOTCAT~\cite{xu2023multimodal}, SurvPath~\cite{jaume2024modeling}, and MoME~\cite{xiong2024mome}.
From the table, it is evident that our method outperforms all others across all five datasets. Specifically, our approach achieves improvements of 0.6\%, 1.4\%, 1.5\%, 2.0\%, and 1.3\% on CESC, LIHC, BRCA, COAD, and KIRC datasets, resulting in an overall 1.7\% boost compared to the runner-up (0.655 \textit{vs.} 0.638). Besides, dual-modal methods show better performance compared to single-modal methods. Our method, driven by descriptive text, effectively leverages triple clinical modal data, leading to the best results.

\subsubsection{Kaplan-Meier Analysis.}
\begin{figure*}[t!]
	\centering
	\includegraphics[width=1\linewidth]{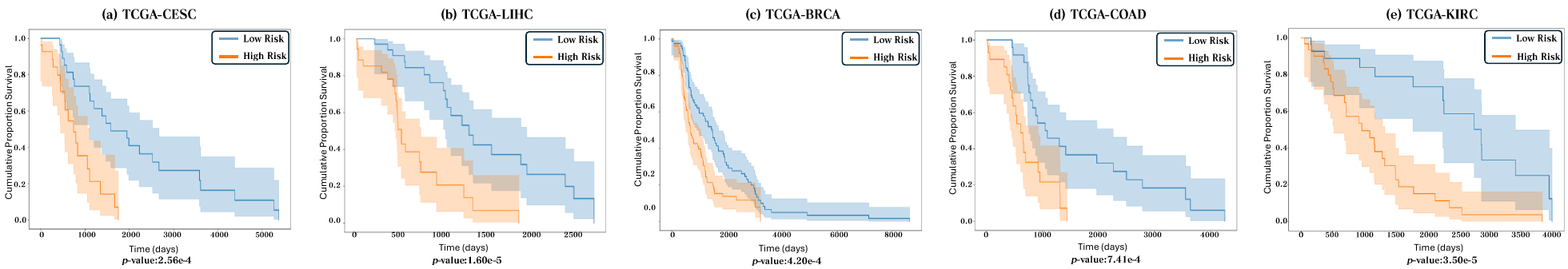}
	\caption{Kaplan-Meier survival curves. The prognostic separation between\colorbox{hr}{\makebox[1.12cm][l]{\textcolor{red_AZUKA}{high-risk}}}~and\colorbox{lr}{\makebox[1.02cm][l]{\textcolor{blue_REI}{low-risk}}} cohorts are stratified by median prognostic scores. The shaded areas denote the confidence intervals. Please zoom in for the best view.}
	\label{fig3}
\end{figure*}
We further conduct statistical analysis based on Kaplan-Meier analysis, and the visualization can be found in Fig.~\ref{fig3}.
The patients are divided into high-risk and low-risk groups based on the cut-off of the predicted median prognostic scores. Our method demonstrates a clear distinction between the two groups, as evidenced by the $p$-values with log-rank test, which are consistently below 0.01 across all five datasets. This indicates a high level of statistical significance and robustness in our method, underscoring the effectiveness of our approach in differentiating between the groups.
	
\subsection{Ablation Studies}

\begin{table}[t]
	\caption{Ablation study of methods on the five datasets. The $^{\dagger}$ denotes using vanilla text prompt, \ie, "\texttt{An H\&E stained image of {\textbf{\{cancer\}}}}".}
	\renewcommand{\arraystretch}{1.1}
	\centering
	\resizebox{1\textwidth}{!}{
		\begin{tabular}{ccccccc}
			\Xhline{3\arrayrulewidth}
			 \multicolumn{1}{c|}{\texttt{CONFIG}} & \multicolumn{1}{c}{\begin{tabular}[c]{@{}c@{}}CESC\end{tabular}} & \multicolumn{1}{c}{\begin{tabular}[c]{@{}c@{}}LIHC\end{tabular}} & \multicolumn{1}{c}{\begin{tabular}[c]{@{}c@{}}BRCA\end{tabular}} & \multicolumn{1}{c}{\begin{tabular}[c]{@{}c@{}}COAD\end{tabular}}& \multicolumn{1}{c}{\begin{tabular}[c]{@{}c@{}}KIRC\end{tabular}} & 
			 \multicolumn{1}{|c}{Overall} \\
			\Xhline{2\arrayrulewidth}
			\multicolumn{1}{c|}{Only Intra-Learning} &$0.610_{\pm 0.034}$&$0.602_{\pm 0.047}$&$0.619_{\pm 0.033}$&$0.601_{\pm 0.026}$&$0.652_{\pm 0.017}$&\multicolumn{1}{|c}{0.617} \\\hline
			 \multicolumn{1}{c|}{Basic} &$0.621_{\pm 0.011}$&$0.611_{\pm 0.024}$&$0.633_{\pm 0.010}$&$0.603_{\pm 0.026}$&$0.656_{\pm 0.034}$&\multicolumn{1}{|c}{0.625} \\
			 \hline
			 \multicolumn{1}{c|} {\texttt{w/o CoC-1} (M1)} 
			 &$0.629_{\pm 0.062}$&$0.616_{\pm 0.043}$&$0.640_{\pm 0.041}$&$0.623_{\pm 0.030}$&$0.690_{\pm 0.046}$&\multicolumn{1}{|c}{0.640} 
			\\
			 \hline
			\multicolumn{1}{c|} {\texttt{w/ CoC-1} (M2)} 
			& $0.638_{\pm 0.021}$& $0.626_{\pm 0.014}$& $0.641_{\pm 0.031}$& $0.626_{\pm 0.034}$& $0.694_{\pm 0.024}$&\multicolumn{1}{|c}{0.645} \\
			 \hline
			 \multicolumn{1}{c|}{\texttt{w/o CoC-3} (M3)} 
			  &$0.613_{\pm 0.037}$ &$0.622_{\pm 0.025}$&$0.620_{\pm 0.051}$&$0.615_{\pm 0.042}$&$0.686_{\pm 0.034}$&\multicolumn{1}{|c}{0.631}\\
			 \hline
			  \multicolumn{1}{c|}{\texttt{w/ CoC-3} (M4)} &$0.624_{\pm 0.030}$&$0.618_{\pm 0.024}$&$0.636_{\pm 0.017}$&$0.627_{\pm 0.049}$&$0.701_{\pm 0.034}$&\multicolumn{1}{|c}{0.641}\\
			 \hline
			 \multicolumn{1}{c|}{+\texttt{CoC-1}~\&~\texttt{CoC-3}($\dagger$M5)}&$0.630_{\pm 0.028}$&$0.620_{\pm 0.037}$&$0.642_{\pm 0.041}$&$0.630_{\pm 0.041}$&$0.697_{\pm 0.034}$&\multicolumn{1}{|c}{0.644} \\
			 \hline
			 \rowcolor{purple_SHINJI!8} 
			  \multicolumn{1}{c|}{+\texttt{CoC-1}~\&~\texttt{CoC-3}(Ours)}&{\textbf{0.643}$_{\pm 0.028}$} &{\textbf{0.630}$_{\pm 0.047}$}&{\textbf{0.654}$_{\pm 0.036}$}&{\textbf{0.638}$_{\pm 0.046}$}&{\textbf{0.709}$_{\pm 0.048}$}& \multicolumn{1}{|c}{{\textbf{0.655}}}  \\
			 \hline
			\Xhline{2\arrayrulewidth}
		\end{tabular}
	}\label{tab2}
\end{table}

We present ablation studies in Tab.~\ref{tab2}. Note that the `Only Intra-Learning' means we only conduct the branch (b) in Fig.~\ref{fig1}. On top of it, the `Basic' merely uses WSIs and CoC-2\&4 in branch (a). And we gradually introduce other modalities to form M1 to M4, and our final method.
\subsubsection{Text Prompt.} In the comparison of using versus not using the CoC-Adapter, we denote them by `w/' and `w/o', respectively. Note we give a linear projection for `w/o' to adjust the shape in order to use AMT. After deploying CoC-Adapter, almost every dataset receives an improvement,  as demonstrated in the overall comparisons between M2 and M1 (0.645~\textit{vs.}~0.640), and M4 and M3 (0.641~\textit{vs.}~0.631). This indicates the effectiveness of our text prompt and adapter design. We also explore the use of chain-of-thought textual descriptions. When deploying the vanilla text prompt for these adapters (M5), performance decreases by 1.1\% compared to our final model. This suggests that the chain-of-thought mechanism is effective for our task.
\subsubsection{Clinical Modal Contribution.} With the introduced AMT, our `Basic' model can interact with the local patches (CoC-2) and global slides (CoC-4) for pathology image, yielding an overall 0.8\% improvement compared to the `Only Intra-Learning'. Based on this `Basic' setting, we can observe that after adding other modalities (M1-M4) in our intra-learning branch (Fig.~\ref{fig1} (a)),  our methods receive gains as well. Moreover, without using methylation data(M1 and M2), our methods still overhead the counterparts (\ie, gene+WSI, methods in Tab.~\ref{table1}). Our final model can leverage triple clinical modalities, producing the best practice. These findings illustrate our AMT can encourage mutual learning among different modalities, and each modality contributes to the final prediction.

%By removing the usage of methylation (M1 and M2), our methods still overhead the counterparts (\ie, gene+WSI).
\section{Conclusion}
In this paper, we propose a Chain-of-Cancer (CoC) framework for survival prediction, based on the pathogenesis of cancer, utilizing three modalities of clinical data and language guidance for the first time.
The core idea of CoC is to innovatively introduce clinical-related descriptions as textual guidance embedded into the clinical features. We propose an Autoregressive Mutual Traction (AMT) module to encourage synergistic learning among different modalities.
Experimental results confirm the effectiveness of our approach, demonstrating its superiority over \textit{ state-of-the-art} methods. Additionally, ablation studies indicate that the proposed designs are effective. 
Our research indicates that by introducing language guidance for multimodal learning, the interpretation can offer significant benefits for feature enhancement.
We hope our study offers valuable insights for future research in multimodal learning for survival prediction.
~\\
~\\
%\noindent\textbf{Acknowledgment. }
%This work is supported by the Guangzhou-HKUST(GZ) xxx.
\bibliographystyle{splncs04.bst}
\bibliography{reference}
\end{document}